\documentclass{article}

\PassOptionsToPackage{numbers, compress}{natbib}
\usepackage[preprint]{neurips_2019}

\usepackage[utf8]{inputenc} 
\usepackage[T1]{fontenc}    
\usepackage{hyperref}       
\usepackage{url}            
\usepackage{booktabs}       
\usepackage{amsfonts}       
\usepackage{nicefrac}       
\usepackage{microtype}      

\usepackage{graphicx, subcaption}
\usepackage[justification=centering]{caption}
\usepackage{adjustbox}
\usepackage{scrextend}
\usepackage{algorithm2e}
\usepackage{mathtools}
\usepackage[inline]{enumitem}
\usepackage{todonotes}

\graphicspath{{Images/}}

\providecommand{\e}[1]{\ensuremath{\times 10^{#1}}} 

\title{Rethinking Full Connectivity in Recurrent Neural Networks}

\author{%
	Matthijs Van Keirsbilck \\
	NVIDIA \\
	Berlin, Germany \\
	\texttt{matthijsv@nvidia.com} \\
	\And
	Alexander Keller \\
	NVIDIA \\
	Berlin, Germany \\
	\texttt{akeller@nvidia.com} \\
	\And
	Xiaodong Yang \\
	NVIDIA \\
	Santa Clara, US \\
	\texttt{xiaodongy@nvidia.com} \\
}%

\begin{document}
	
	\newpage
	
	\maketitle
	
	\begin{abstract}
		Recurrent neural networks (RNNs) are omnipresent in sequence modeling tasks. Practical models usually consist of several layers of hundreds or thousands of neurons which are fully connected. This places a heavy computational and memory burden on hardware, restricting adoption in practical low-cost and low-power devices. Compared to fully convolutional models, the costly sequential operation of RNNs severely hinders performance on parallel hardware.
		This paper challenges the convention of full connectivity in RNNs. We study structurally sparse RNNs, showing that they are well suited for acceleration on parallel hardware, with a greatly reduced cost of the recurrent operations as well as orders of magnitude less recurrent weights.
		Extensive experiments on challenging tasks ranging from language modeling and speech recognition to video action recognition reveal that structurally sparse RNNs achieve competitive performance as compared to fully-connected networks.
		This allows for using large sparse RNNs for a wide range of real-world tasks that previously were too costly with fully connected networks.
	\end{abstract}

	\section{Introduction}
	
	Fully connected neural networks are known to have many redundant weights \cite{deepcompression,mocanu2018scalable,narang2017block}, which makes training challenging and inference costly. There is an need for increasingly large networks \cite{openaiCompute2018}, but the compute and memory cost increases quadratically with network size if fully connected, which makes the cost of increasing network size prohibitive. %
	
	Large RNNs are increasingly being replaced by convolutional sequence-to-sequence networks because of the RNN's huge number of trainable weights. This causes overfitting and training instability, and limits the potential of parallellization \cite{bai2018,gehring2017convolutional}.
	Techniques like pruning or factorization have been proposed to reduce the inference cost of these networks but often increase the training time and complexity and introduce additional hyperparameters. In addition, achieving real-world speedups remains difficult. 
	
	In contrast to fully-connected RNNs, neurons in the human brain and other real-world resource-constrained systems are not fully connected at all, with the majority of the connections being local \cite{chechikNeuralPlasticity, mocanu2018scalable, watts1998collective}. 
	We review structurally sparse, locally recurrent neural network architectures, and discuss implications for hardware acceleration.
	
	Compared to unstructured sparsity, for example by pruning, structural sparsity also reduces the number of weights and computations, but it does so in a way much more suitable for parallel hardware acceleration. 
	The sparse recurrent connections provide the additional advantage of reduced communication between neurons, enabling more parallelism.
	In addition, learning about ways to simplify the network architecture grants insight into the functioning of these networks.

	\section{Related Work}\label{s:relatedWork}
	
	It has been shown that a large sparse network performs better than a small dense network \cite{han2015learning,kalchbrenner2018,narang2017block}. \citet{han2015deep} combined iterative pruning-retraining, quantization and encoding to achieve 50$\times$ model compression on AlexNet. 
	Achieving real-world speedups remains difficult, as the technique results in unstructured sparsity and irregular memory access patterns.
	
	Structured pruning modifies the loss function to create more hardware-friendly memory access patterns, resulting in block-sparse weight matrices, for example. \citet{narang2017block} achieved the same performance as a full LSTM using 10 times less weights and a similar speedup. An alternative approach is to factorize the weight matrices \cite{KuchaievG17, sainath2013low}.
	Both techniques are effective for improving speed on current hardware, but offer little architectural insight into the reasons the resulting sparse networks are so good. In addition, training is still fairly complex and expensive, which limits the practical use of large networks.
	
	\paragraph{Self-Connected Recurrent Neural Networks.} The sequential modeling capabilities of RNNs stem from the feedback connections. The most direct feedback connections are self-connections of neurons through which neurons can access information about their own past. This idea leads to the DiagonalRNN architecture, where neurons only have self-connections, and the recurrent weight matrix is a diagonal matrix (in fact, simply a vector).
	
	DiagonalRNNs have been used successfully in systems control since the '90s  \cite{JiDiagRNN, ku1995diagonal, sivakumar1999online}. There are no recurrent connections between neurons in the same layer except for self-connections. This reduces the recurrent transformation to an $O(n)$ element-wise vector-vector multiplication instead of an $O(n^2)$ matrix-vector multiplication. The number of recurrent weights is reduced correspondingly.
	
	Different variations of this architecture have been studied, varying the location of the feedback and the type of recurrence transformation (in general this feedback can be considered an IIR filter \cite{back1991fir}).
	Alternatively, FIR-type recurrences are possible, which are closely related to convolutional sequence-to-sequence models (\citet{van2016wavenet}). \citet{tsoi1994locally} provide an overview.
	
	The DiagonalRNN idea was recently rediscovered and applied to LSTMs and GRUs by \citet{diagRNNmusic}. It was also explored by \citet{indrnn} for deep vanilla RNN networks with skip connections between them. Both articles show performance similar to fully connected LSTMs, however, at much lower cost.  \citet{mikolovSCRNN} use the DiagonalRNN with fixed recurrent weights close to unity as contextual units in addition to a fully-connected vanilla RNN, obtaining performance close to LSTMs yet with a much simpler architecture.
	
	One disadvantage of DiagonalRNN is that neurons can no longer see the hidden states of other neurons in the same layer, placing constraints on the model's temporal dynamics. However, even with this constraint the models perform well on many challenging tasks as shown in our experiments. 
	
	\paragraph{Locally Recurrent Neural Networks.} To combine the advantages of parallelism, structured memory access patterns, linear relation between number of neurons and the computational cost and memory footprint, as well as the temporal modeling capacity of the fully-connected models, locally connected networks are a promising alternative to self-recurrent networks.
	This approach is inspired by both biological neural networks and other real-world networked systems, where each connection incurs a cost. Such systems exhibit local connectivity and operate extremely well even under strict connectivity constraints \cite{chechikNeuralPlasticity}. 
	
	\citet{chan1993ring} propose locally connected RNNs: Within one layer each neuron is only connected to itself and a few neighbours instead of connecting to all other neurons or only to itself (see Figure~\ref{f:BandRNN}). This allows for greater temporal modeling capability, while reducing the size of the hidden weight matrix from $N\times N$ to $N\times C$, with $C$ the number of connections per neuron being much smaller than $N$. By varying the number of neighbours a neuron is connected to, a trade-off between the complexity of temporal dynamics and the computational complexity can be achieved.
	
	The limited local connectivity also allows for better control of network stability during training, which is a challenge with fully connected RNNs. The constraint imposed by local connectivity can act as a regularizer, making learning easier and reducing overfitting \cite{sivakumar1999online}.
	
	A network with local recurrent connections only results in a band-diagonal recurrent weight matrix $W_{rec}$. There exist fast linear algebra routines for computing banded matrix-vector products, so this recurrent transformation is straightforward to accelerate on current hardware.
	
	A variant of the locally-connected RNN is the ``grouped'' RNN \cite{demeester2018, KuchaievG17, zhu2017parallelcellrnn}, where a single layer is replaced by several smaller ones in parallel. Neurons are partitioned in several groups, with full connectivity within each group. Groups are not connected to each other, but their outputs are concatenated when passed to the next layer. This can be seen as a generalization of DiagonalRNNs, where the independent units are now groups of several neurons instead of single neurons.
	Interestingly, this idea is also very similar to the idea of grouped convolutions, which is gaining traction in state-of-the-art convolutional neural networks \cite{zhang2017shufflenet}.
	
	\citet{gray2017blocksparsegpu} introduced fast GPU kernels for computations with block-sparse matrices, and demonstrated good performance of block-sparse and small-world RNNs on large-scale sentiment analysis and speech recognition tasks.
	
	Both \citet{sru} and \citet{qrnn} propose alternative gated RNNs with a very efficient element-wise recurrent transformation. The gates are computed based only on the current input, so these operations can be easily executed in parallel across time steps. The recurrence is very fast as compared to full matrix multiplications at every time step.
	In several tasks this architecture has been shown to perform well, however the lack of a hidden state may limit the model capacity \cite{merityLargeLM2018}.

	\section{Structurally Sparse RNNs for Hardware Acceleration}\label{s:locallyRecurrentRNNs}
	
\begin{figure*}
		\centering
		\begin{subfigure}{.26\linewidth}
			\includegraphics[width=0.9\linewidth]{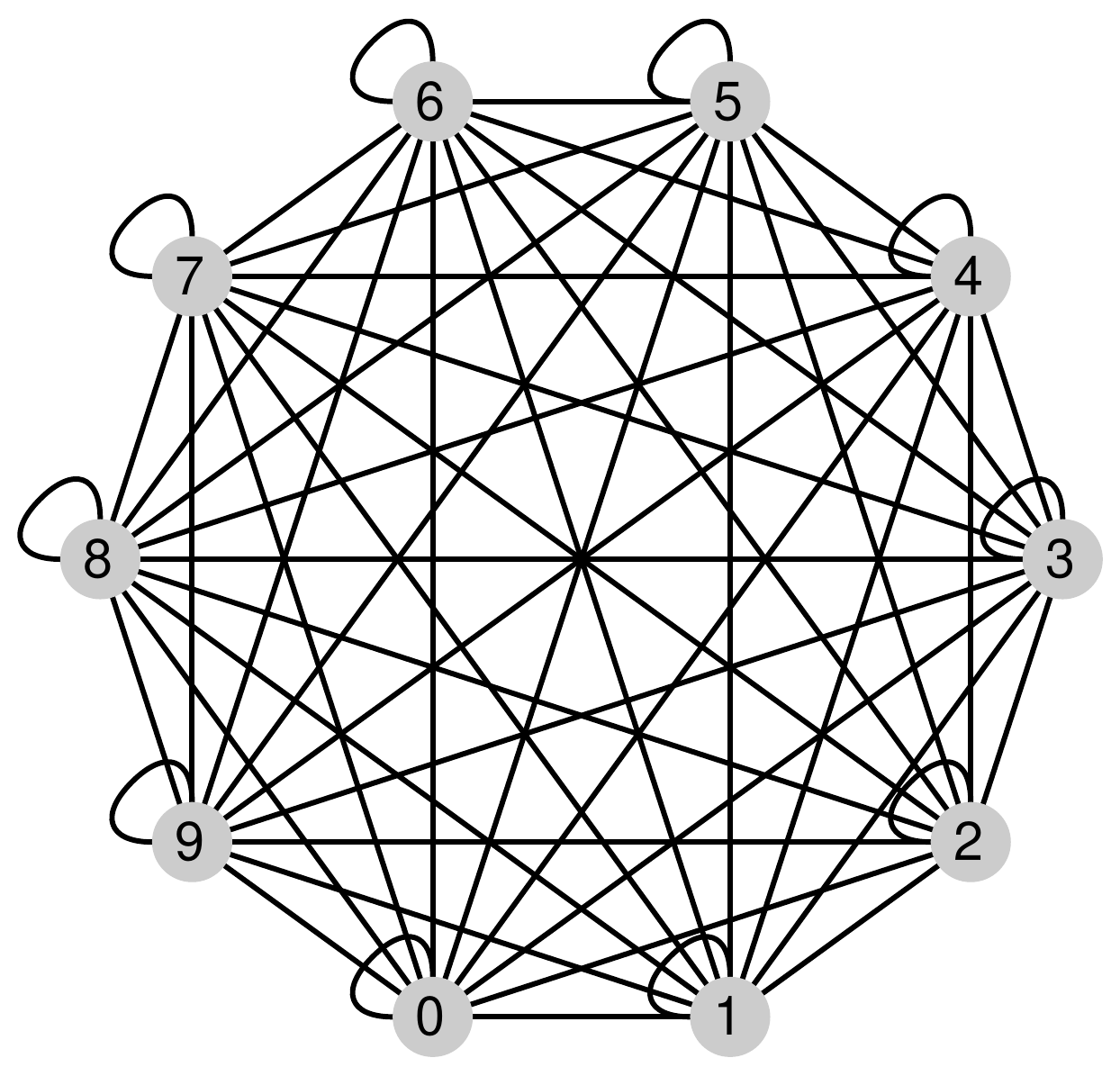}%
			\caption{Full RNN: connections to all neurons in the layer}
			\label{f:FullRNN}
		\end{subfigure}
		\begin{subfigure}{.26\linewidth}
			\includegraphics[width=0.9\linewidth]{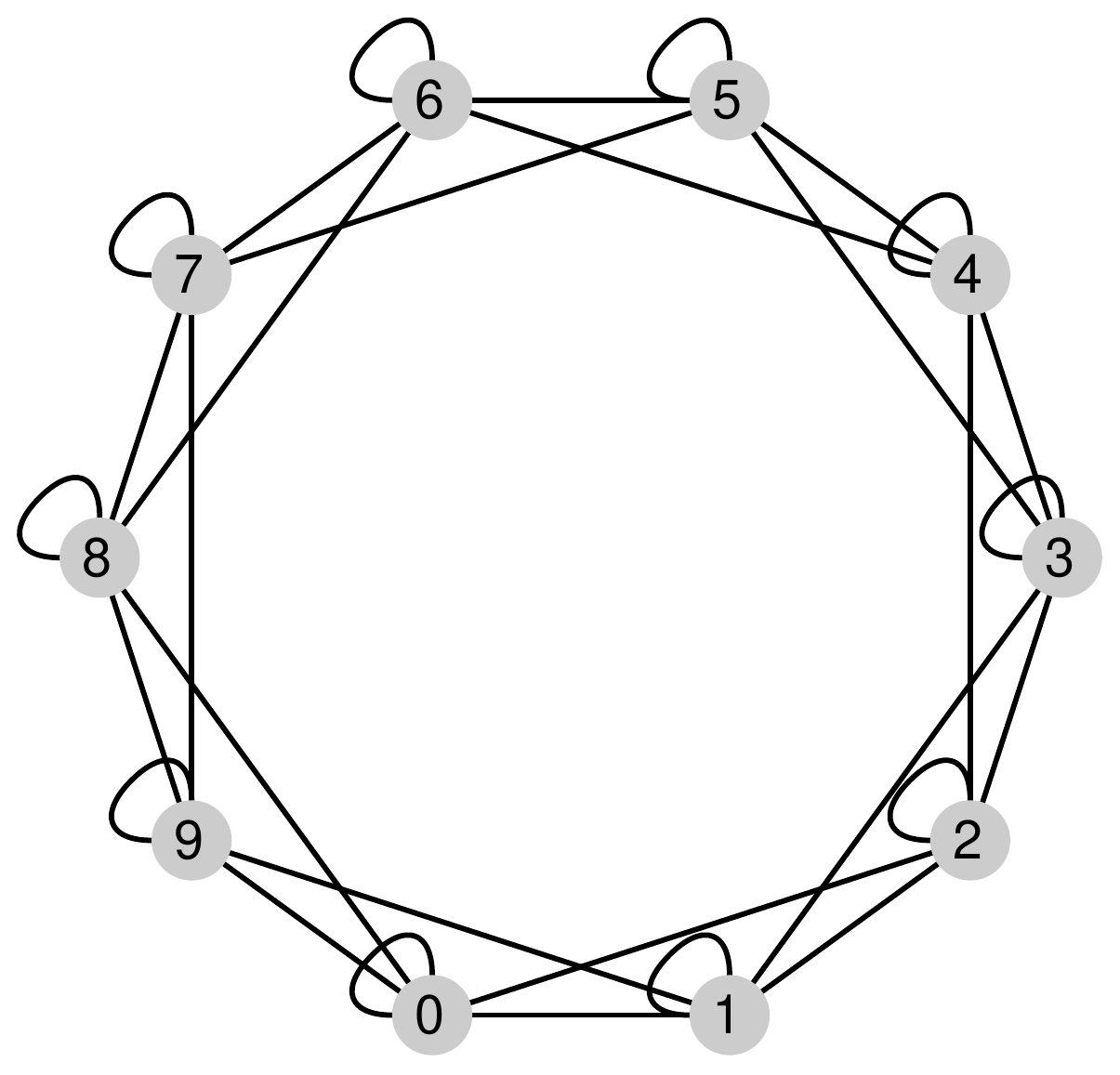}%
			\caption{BandRNN: connections to some neighboring neurons plus self-connection}
			\label{f:BandRNN}
		\end{subfigure}
		\begin{subfigure}{.26\linewidth}
			\includegraphics[width=0.9\linewidth]{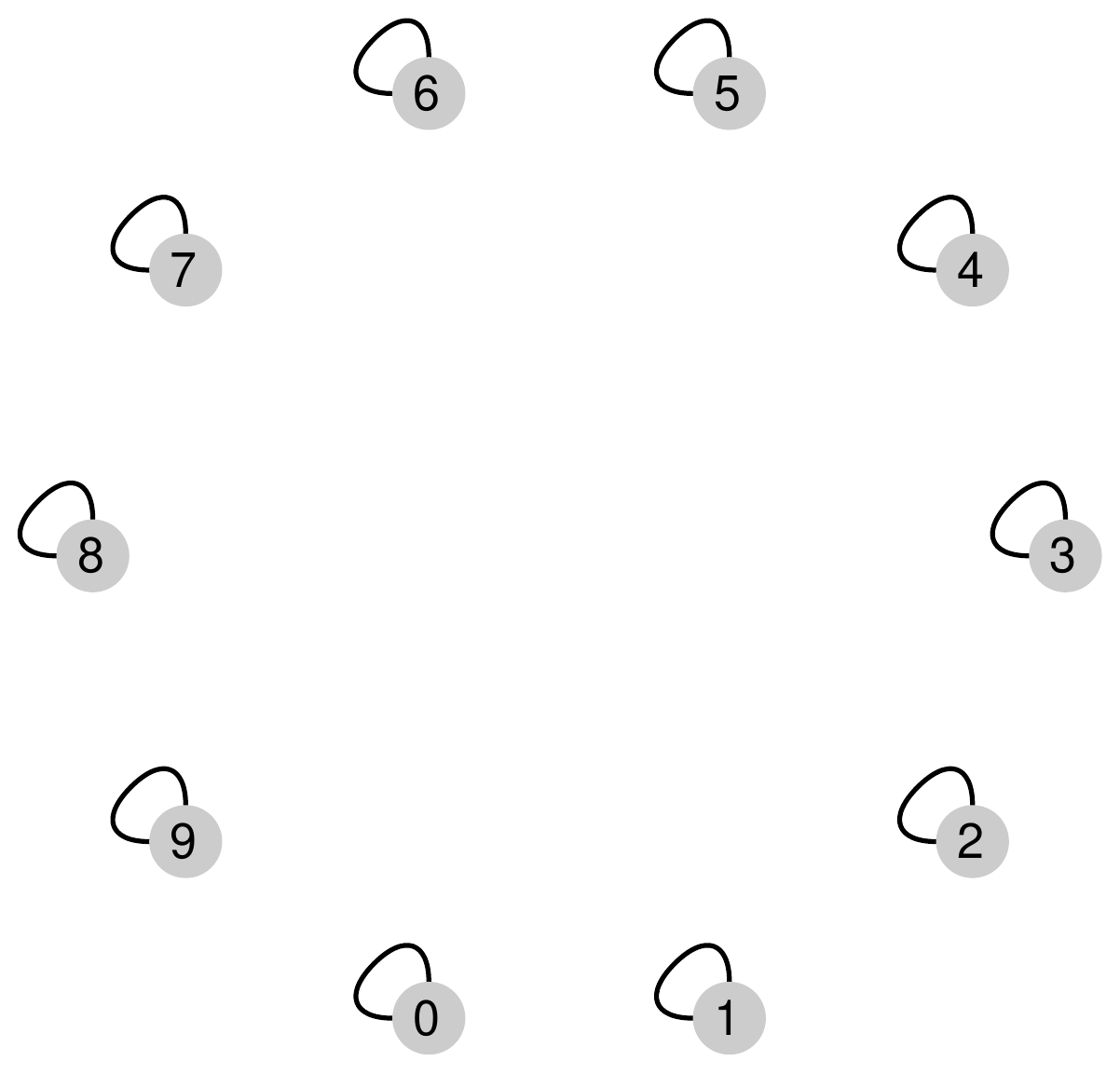}%
			\caption{DiagonalRNN: self-connections only}
			\label{f:DiagonalRNN}
		\end{subfigure}
		\caption{Recurrent connectivity for various RNN types} 
		\label{f:rnnConnectivity}
	\end{figure*}

\begin{figure*}
		\centering
		\begin{subfigure}{.26\linewidth}
			\includegraphics[width=\linewidth]{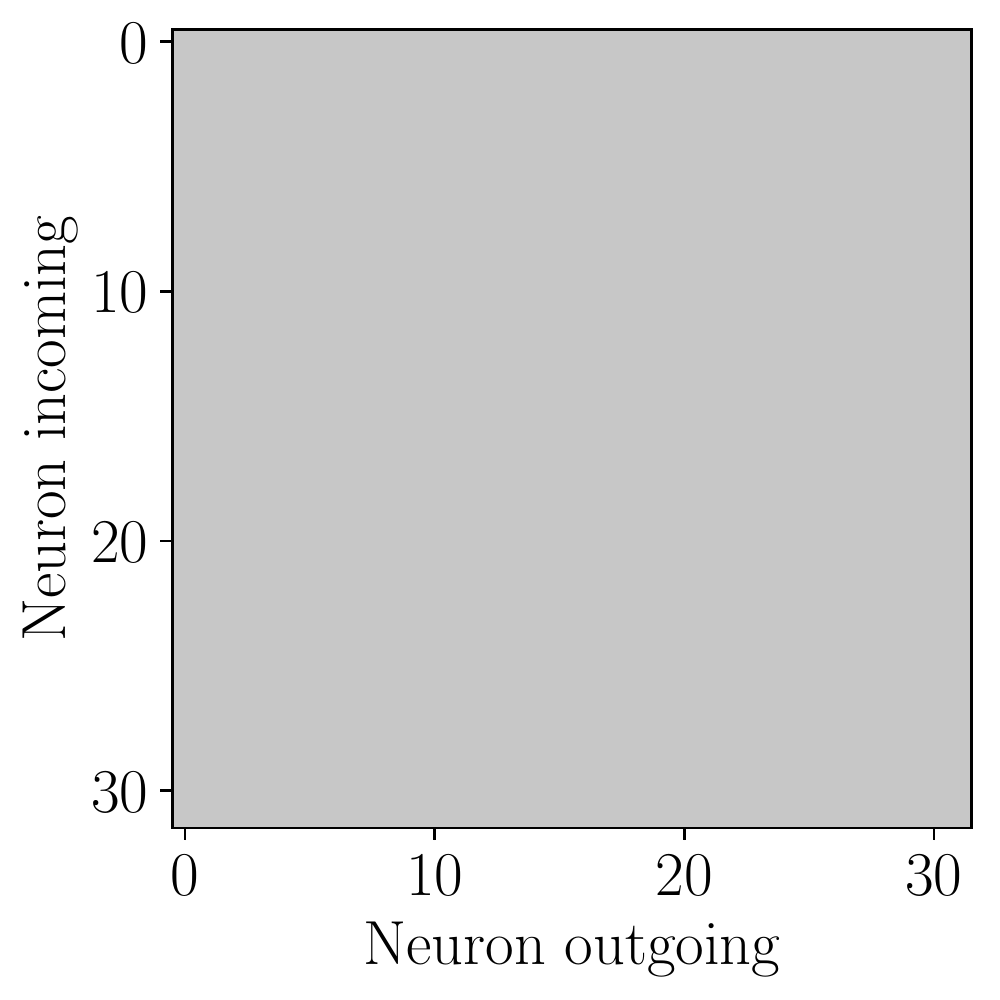}%
			\caption{Full RNN}\label{f:WrecFull}
		\end{subfigure}
		\begin{subfigure}{.26\linewidth}
			\includegraphics[width=\linewidth]{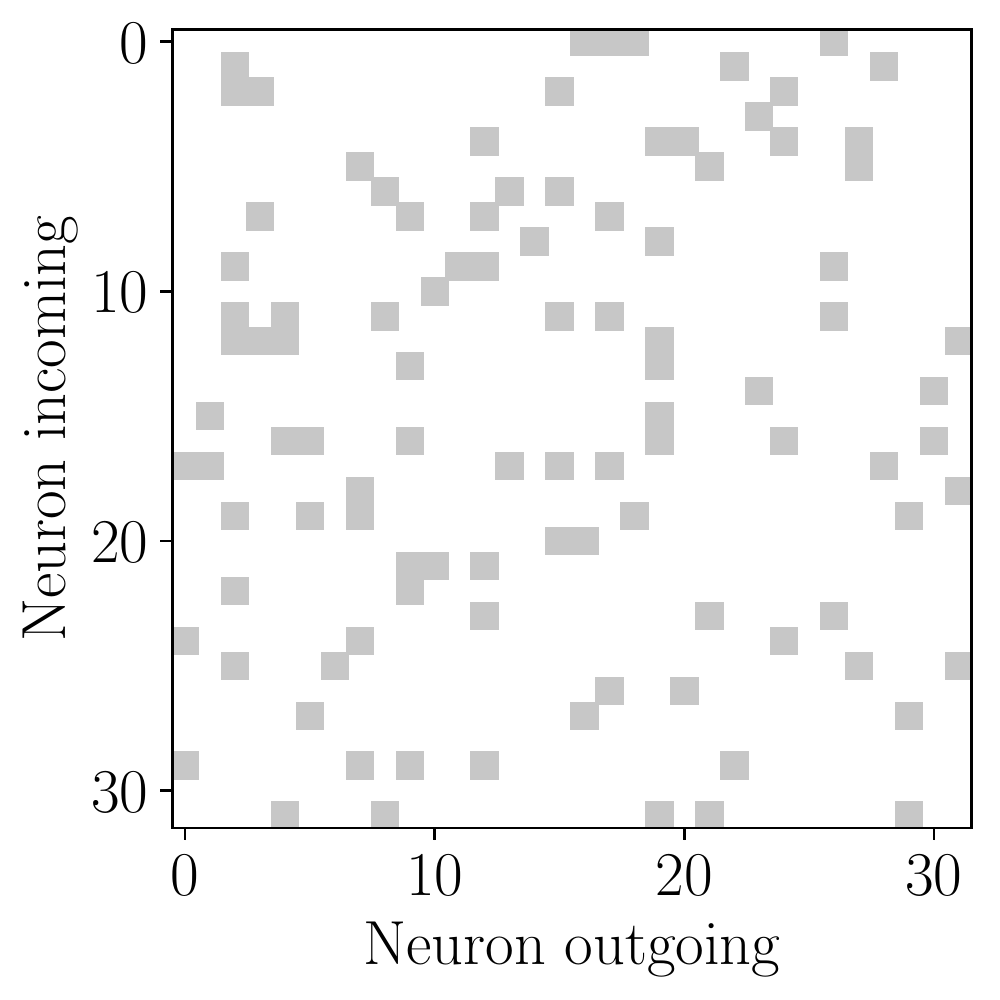}
			\caption{Sparse RNN}\label{f:WrecSparseUnstructured}
		\end{subfigure}
		\begin{subfigure}{.26\linewidth}
			\includegraphics[width=\linewidth]{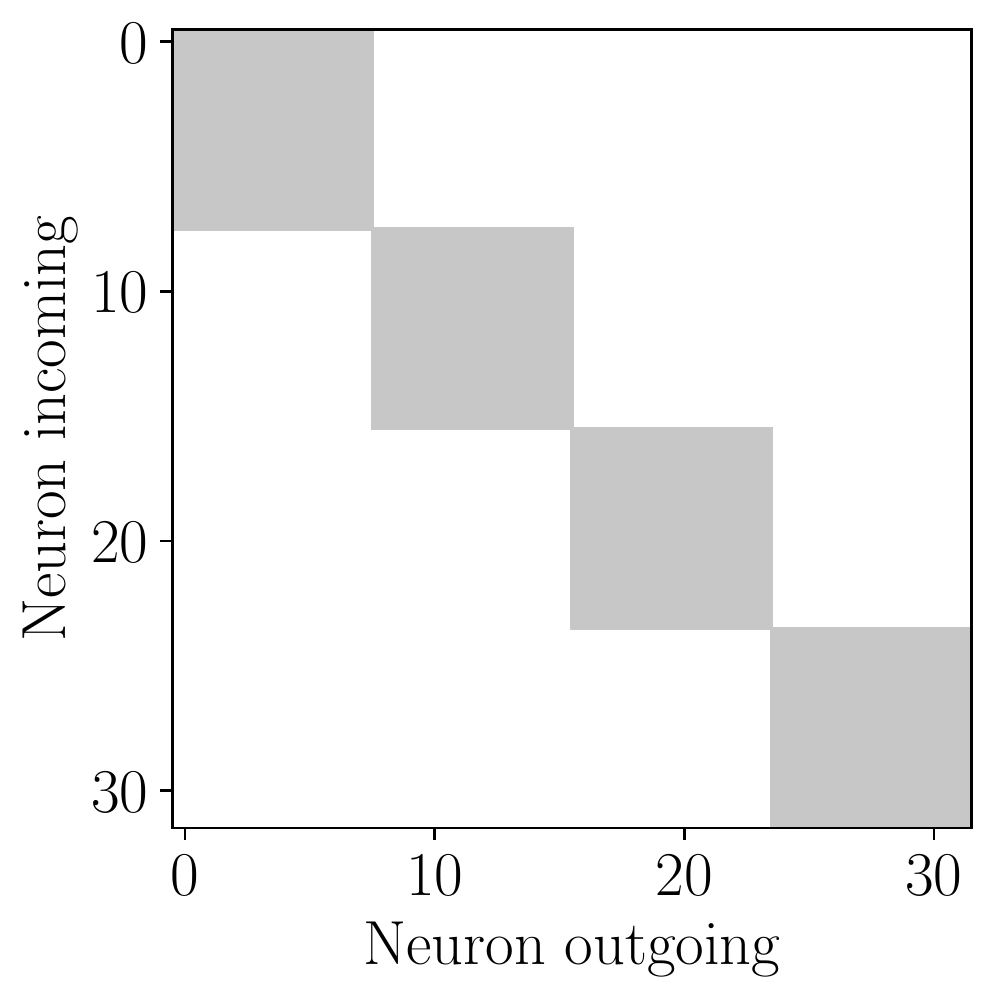}
			\caption{GroupRNN, $G=4$}\label{f:WrecGrouped}
		\end{subfigure}
		\\
		\vspace{0.3cm}
		\begin{subfigure}{.26\linewidth}
			\includegraphics[width=\linewidth]{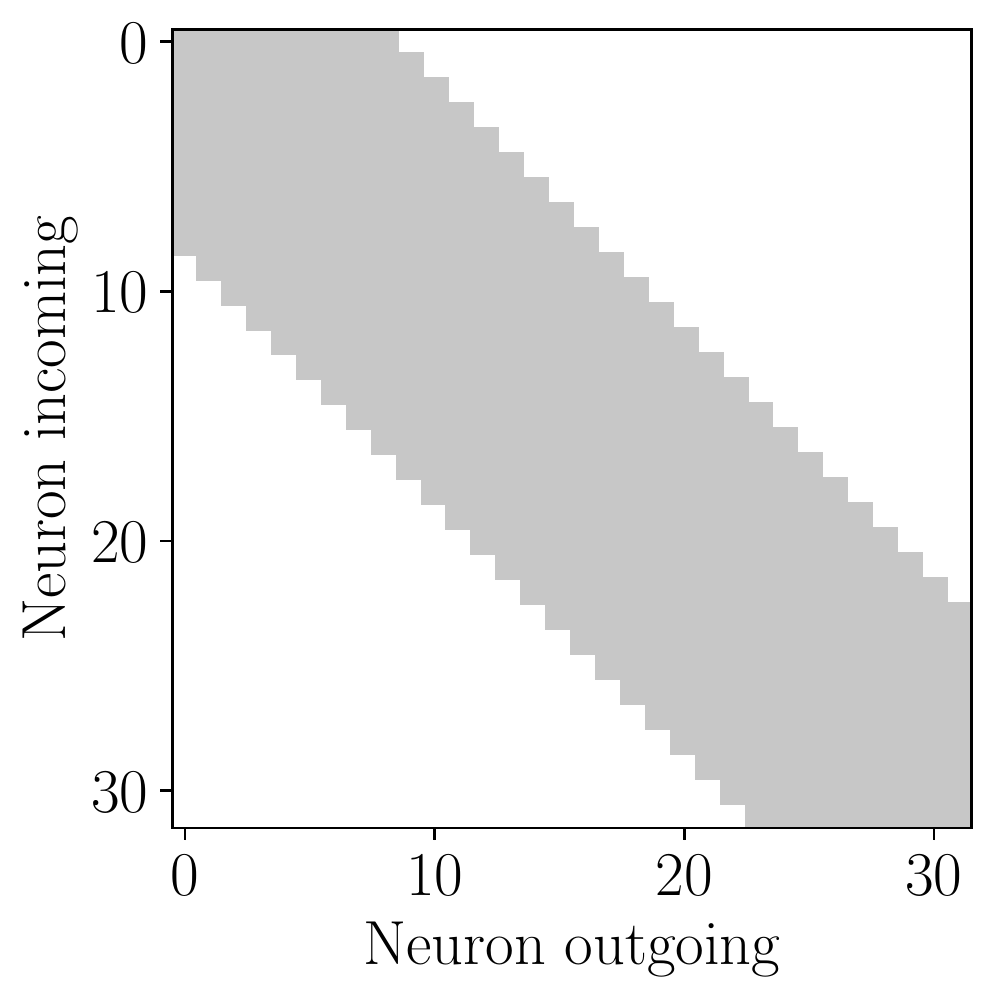}
			\caption{BandRNN, $C=17$}\label{f:WrecBand17}
		\end{subfigure}
		\begin{subfigure}{.26\linewidth}
			\includegraphics[width=\linewidth]{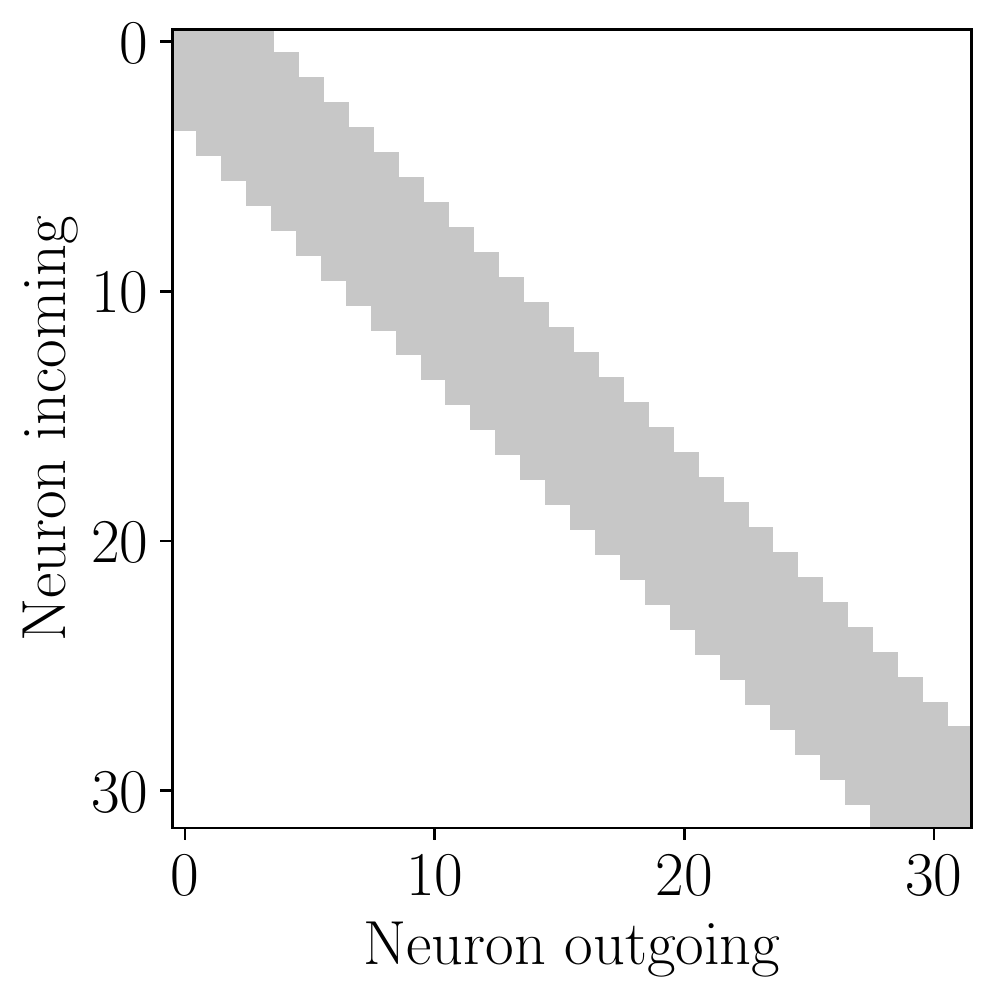}
			\caption{BandRNN, $C=7$}\label{f:WrecBand7}
		\end{subfigure}
		\begin{subfigure}{.26\linewidth}
			\includegraphics[width=\linewidth]{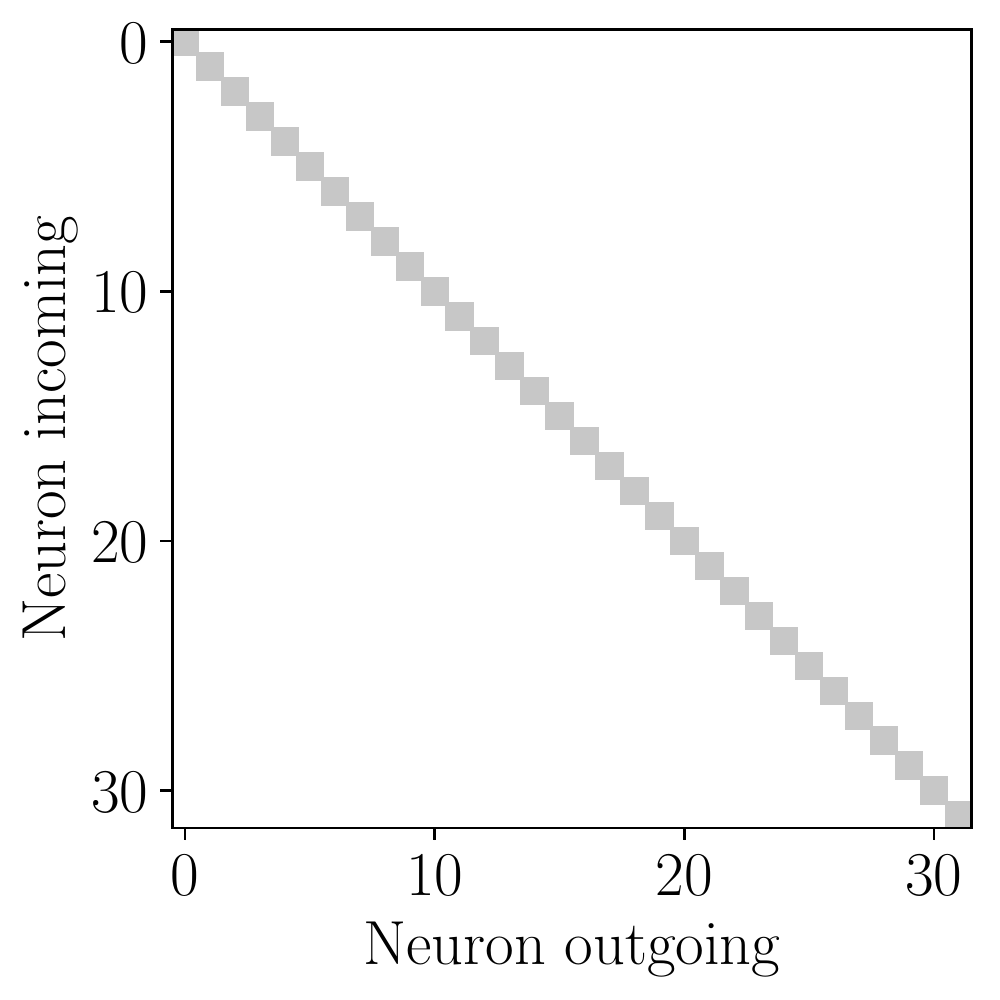}
			\caption{DiagonalRNN, $C=1$}\label{f:WrecDiagonal}
		\end{subfigure}
		
		\caption{Recurrent weight matrices for various RNN types. Grey indicates a nonzero weight, while white indicates a zero weight (\textit{i.e.} there's no connection)}
		\label{f:Wrecs}
	\end{figure*}

	As described in Section~\ref{s:relatedWork}, there are a variety of RNN architectures that make use of local (as opposed to global) connectivity. Table~\ref{t:sparsernn-nbparams} lists an overview of the architectures, together with the complexity of the recurrent weight matrix. Figure~\ref{f:rnnConnectivity} shows the connectivity patterns, while Figure~\ref{f:Wrecs} shows a visual representation of these sparse weight matrices.
	
	In unstructured sparse networks (see Figure~\ref{f:WrecSparseUnstructured}), non-zero weights are spread far apart in an irregular pattern with lots of zeros in between. When loading the weight matrices from memory, these zeros have to be loaded as well, and cause higher latency as well as high energy consumption \cite{deepcompression}.
	
	The structurally sparse matrices of GroupRNN/BandRNN/DiagonalRNN have large advantages with regards to memory access patterns, as their non-zero weights are all located close together in memory. In addition, there are some straightforward parallellization opportunities.
	The following paragraphs provide an overview of the benefits for hardware acceleration of various RNN architectures.
	
	\begin{description}
		\item[Full RNN:]
		in a standard fully-connected recurrent neural network all neurons have recurrent connections to all other neurons. The connectivity pattern is visualized in Figure~\ref{f:FullRNN}, and the recurrent weight matrix in Figure~\ref{f:WrecFull}. There is no opportunity for parallellization across neurons, and the recurrent transformation requires a full matrix multiplication.
		
		\item[GroupRNN:]
		here, neurons are divided in $G$ groups, with full connectivity within each group, but no connections between groups. The recurrent weight matrix structure is shown in Figure~\ref{f:WrecGrouped}.
		This allows for complex temporal dynamics and specialization across groups, while the recurrent computation is much reduced due to the lower number of weights. Each neuron group operates independently of the other groups in the same layer, so there is an opportunity for parallellization as well.
		
		\item[BandRNN:]
		this is a locally connected RNN, where neurons have self-connections in addition to connections to their immediate neighbours. This creates a sparse band-diagonal recurrent weight matrix, which can be exploited for efficient computation. There is a \textit{linear} relation between the number of recurrent weights and the number of neurons. The recurrent weight matrix is shown in Figure~\ref{f:WrecBand17} and Figure~\ref{f:WrecBand7} for several band widths. The connectivity pattern is visualized in Figure~\ref{f:BandRNN}.
		
		\item[DiagonalRNN:]
		this is an extreme form of BandRNN with zero connections to neighbours, or GroupRNN with $G = N_h$. This means every neuron is completely independent of the other neurons in the same layer, and can be computed in parallel. There are \textit{only $N_h$ recurrent weights}, and the recurrent transformation consists of \textit{elementwise operations only}. The connectivity pattern is visualized in Figure~\ref{f:DiagonalRNN}.
	\end{description}
	
	\begin{table}
		\caption{The number of recurrent weights $N_{rec}$ for a layer of $N_h$ neurons, depending on the network connectivity. The number of recurrent weights for DiagonalRNN and BandRNN is \textit{linear} in the number of neurons instead of quadratic}
		\label{t:sparsernn-nbparams}
		\centering
		\begin{tabular}{lc}
			\toprule
			\textbf{RNN Cell type} & \textbf{\# Recurrent Weights} \\
			\midrule
			Full RNN  	& $N_h \cdot N_h$ 		\\ 
			GroupRNN  	& $N_h \cdot N_h / G$ 	\\ 
			DiagonalRNN   	& $N_h$	 			\\ 
			BandRNN 	& $N_h \cdot C$  		\\ 
			\bottomrule
		\end{tabular}
	\end{table}
	
	\subsection{DiagonalRNNs: high speed and low memory footprint}\label{s:speedupOpportunity}
	
	Table~\ref{t:diagrnn-speed} compares the speed and cost of full RNNs with DiagonalRNNs. %
	DiagonalRNNs provide large speedup opportunities through 
	\begin{enumerate}
		\item vastly reduced computation in the (sequential) recurrent operation,
		\item parallellization across neurons,
		\item much lower memory footprint.
	\end{enumerate}
	
	\begin{table}[ht]
		\caption{Speed experiment in PyTorch, comparing cuDNN RNN, and DiagonalRNN with custom CUDA kernel. 
			DiagonalRNN does away with almost all of the recurrent weights, and achieves \textbf{large speedups of 9x} at inference compared to Full RNN.}
		\label{t:diagrnn-speed}
		\centering
		\begin{tabular}{lcccc}
			\toprule
			\textbf{Model} & Train time/batch &  Infer time/batch & Model size & \# Recurrent \\
			$L=1, N_h=512$ & &  &  & Weights \\ 
			\midrule
			cuDNN RNN 		    & 43 ms     & 8.2 ms 	& 47.5 MB   & 262.144  \\ 
			CUDA DiagonalRNN	& 23 ms     & 0.9 ms 	& 25.1 MB   & 512 \\
			\bottomrule
		\end{tabular}
	\end{table}
	
	In DiagonalRNNs, the recurrent computation is much simpler, as it only requires elementwise operations instead of full matrix multiplications. This is of great benefit as the recurrent transformation is the sequential part of the RNN operation, and usually the performance bottleneck. 
	Secondly, every neuron in a layer is completely independent of every other neuron in that layer, so it's possible to parallelize across neurons. This creates an additional level of parallelism, making them very well suited for execution on parallel hardware.
	
	Due to the much lower cost of the sequential recurrent computation, plus the parallellization across neurons, \textbf{DiagonalRNNs are 9x faster than full RNN for inference}. 
	
	In Deep Learning hardware accelerators, the cost of loading weights from memory accounts for a large portion of the total energy cost \cite{deepcompression}.  With DiagonalRNNs, the storage cost for the recurrent weights is $O(N_h)$ instead of $O(N_h^2)$, \textit{i.e.} linear instead of quadratic in the number of neurons. 
	
	This brings the number of recurrent weights to almost zero compared to fully connected networks (see Table~\ref{t:diagrnn-speed}.
	The much smaller memory footprint allows keeping the model weights fully in registers, as proposed by \citet{zhu2018persistent}, which allows for further large increases in efficiency, reduced latency and higher performance.
	
	\subsection{PreRNN with BandRNN}

	RNNs are widely applied to capture motion dynamics and to provide temporal contexts in video understanding.
	Since the processing unit of a video is an image frame or a short video snippet, CNNs pre-trained on large-scale image or video data sets are usually used as backbone networks, which RNNs are then built upon~\cite{dynamic-facial, multilayer}.
	
	Instead of simply stacking a CNN and a full RNN and training the whole network from scratch, PreRNN has recently been proposed in~\cite{prernn} to convert convolutional layers into recurrent layers.
	PreRNN employs one or more pre-trained feedforward layers of a CNN as the input-to-hidden transformations to the RNN and therefore avoids the default input-to-hidden transformation, which uses a matrix multiplication with $N_h \times N_{inputs}$ weights of each gate. 
	The only new weights introduced by PreRNN are then the $N_h \times N_h$ hidden-to-hidden weights for each gate. ~\cite{prernn} also propose the 'PreRNN-SIH' architecture to further simplify gating functions by binding all gates to the same single input-to-hidden transformation. 
	This technique allows for large weight savings, faster training, and in many cases better performance. 
	
	\textbf{PreRNN and BandRNN are highly complementary} as the former significantly simplifies the input-to-hidden transformations and the latter largely reduces the hidden-to-hidden weights. 
	Therefore, the combination of PreRNN and BandRNN leads to a very efficient, high-performance video action recognition model, as demonstrated in Section~\ref{s:experiments}.
	
	\section{Experiments}\label{s:experiments}
	
	Networks are in the experiments are specified by the recurrent cell type (RNN or LSTM), the number of layers $L_h$, and the number of neurons per layer: $N_{h}$. BandRNN networks are additionally specified by the number of recurrent weights per neuron, $C$.
	For example: BandRNN, $1\times512, C=11$ is an RNN with 1 layer of 512 neurons, with 11 connections per neuron (1 self-connection, 5 to neighbours on the left and 5 to neighbours on the right).
	
	\subsection{Language Modeling: PTB} \label{s:experimentsPTB}
	PTB is a well-known language modeling data set \cite{ptb}.
	We perform word-level language modeling on the PTB data set, with a vocabulary size of 10000, embedding size of 512 and tied weights for encoder/decoder, using the Adam optimizer and a learning rate of 0.001. The remaining training setup is according to \citet{merityRegularizing}, using training code at \footnote{\url{https://github.com/salesforce/awd-lstm-lm}}.
	
	Table~\ref{t:ptb} shows the perplexity results for different models. DiagonalLSTM achieve test perplexity very close to the fully connected LSTM, and both GroupLSTM and BandLSTM improve the perplexity.
	When increasing the band, performance actually decreases, so it appears that limiting the recurrent connectivity has a regularizing effect, which is especially important in the PTB benchmark.
	
	\begin{table}
		\caption{PTB Language Modeling. DiagonalLSTM performs equivalently to full LSTM, while requiring about $1000\times$ less recurrent weights. BandLSTM improves on full LSTM, and requires $50\times$ less recurrent weights.}
		\label{t:ptb}
		\centering
		\begin{tabular}{lccc}
			\toprule
			\textbf{Model} & \textbf{Test Perplexity} &  \multicolumn{2}{c}{\textbf{RNN Weights}} \\
			\cmidrule(r){3-4}
			$L=3, N_h=1150$ & (lower is better) & Total & Recurrent \\ 
			\midrule
			Full LSTM 			    & 65.09             & 2.02\e{7} & 1.12\e{7} \\ 
			GroupLSTM, $G=4$ 	    & 63.81             & 1.17\e{7} & 2.80\e{6} \\
			DiagonalLSTM, $C=1$		& 65.02             & 8.98\e{6} & 1.08\e{4} \\ 
			BandLSTM, $C=23$ 	    & 63.70             & 9.21\e{6} & 2.43\e{5} \\ 
			BandLSTM, $C=105$ 	    & 63.85             & 1.01\e{7} & 1.11\e{6} \\
			BandLSTM, $C=460$       & 64.06             & 1.30\e{7} & 4.05\e{6} \\ 
			\bottomrule
		\end{tabular}
	\end{table}
	
	\subsection{Phoneme Recognition: TIMIT}\label{s:experimentsTIMIT}
	
	TIMIT is a well-known speech recognition data set containing frame-level phonetic annotations labeled by experts \citet{timit}. It contains 5.4 hours of audio data from 630 different speakers of eight American English dialects, with rich and varied phonetic content \cite{timit}.
	
	For the task of phoneme recognition, the goal is to predict the sub-word phoneme corresponding to every frame of audio. This is a more low-level task than speech recognition, and considering the small size of the data set a good task for fast benchmarking of networks.
	We use 12 MFCC coefficients plus energy with first and second derivatives for a total of 39 input features. Labels are the TIMIT phoneme labels, collapsed to the reduced set of 39 phonemes proposed by \citet{lee1989speaker}. The initial learning rate is 0.001, with a decay factor of 0.5 every time the accuracy doesn't improve for 3 subsequent epochs. We use the Adam optimizer. Networks are trained for 25 epochs.
	
	Table~\ref{t:timit} shows an overview of the experiments. The sparse networks achieve encouraging results, with DiagonalLSTM performing slightly worse than the full LSTM, but while requiring $500 \times$ less recurrent weights. BandLSTM performs significantly better than the fully connected network, while still requiring $5 \times$ less recurrent weights.
	
	\begin{table}
		\caption{Phoneme recognition on TIMIT. Structurally sparse networks perform close to or better than the fully connected network, while requiring much less weights.}
		\label{t:timit}
		\centering
		\begin{tabular}{lccc}
			\toprule
			\textbf{Model} & \textbf{Test Accuracy} &  \multicolumn{2}{c}{\textbf{RNN Weights}} \\
			\cmidrule(r){3-4}
			$L=1, N_h=512$ & (higher is better) & Total & Recurrent \\ 
			\midrule
			Full LSTM 		    & 78.3 			& 1.13\e{6} & 1.05\e{6} \\ 
			DiagonalLSTM		& 77.6 			& 8.19\e{4} & 2.05\e{3} \\
			BandLSTM, $C=103$   & 78.8          & 2.81\e{5} & 2.11\e{5} \\
			\bottomrule
		\end{tabular}
	\end{table}
	
	\subsection{Speech Recognition: VCTK}\label{s:experimentsVCTK}
	VCTK is a standard speech recognition data set consisting of 44 hours of audio from 109 native English speakers, reading sentences selected from a newspaper \citet{vctk}. It is a large-scale, open vocabulary problem and a challenging task for speech recognition systems \cite{vctk}.
	
	Models are trained to perform character-level speech recognition. The network architecture and training setup is very similar to Deepspeech2 (\citet{deepspeech2}), consisting of 2 convolutional layers, followed by 5 recurrent layers, with the CTC loss function. There is no language model used, and greedy decoding is used. We use a 90-10 train/test split, selecting randomly. The optimizer is SGD with a momentum of 0.9, with an initial learning rate of 0.005, decreasing by factor 0.9 every epoch, and we train for 20 epochs. We use small tempo and gain for data augmentation. Training code is based on \footnote{\url{https://github.com/SeanNaren/deepspeech.pytorch}}. Further details can be found there.
	
\begin{figure}
		\centering
		\includegraphics[width=0.65\linewidth]{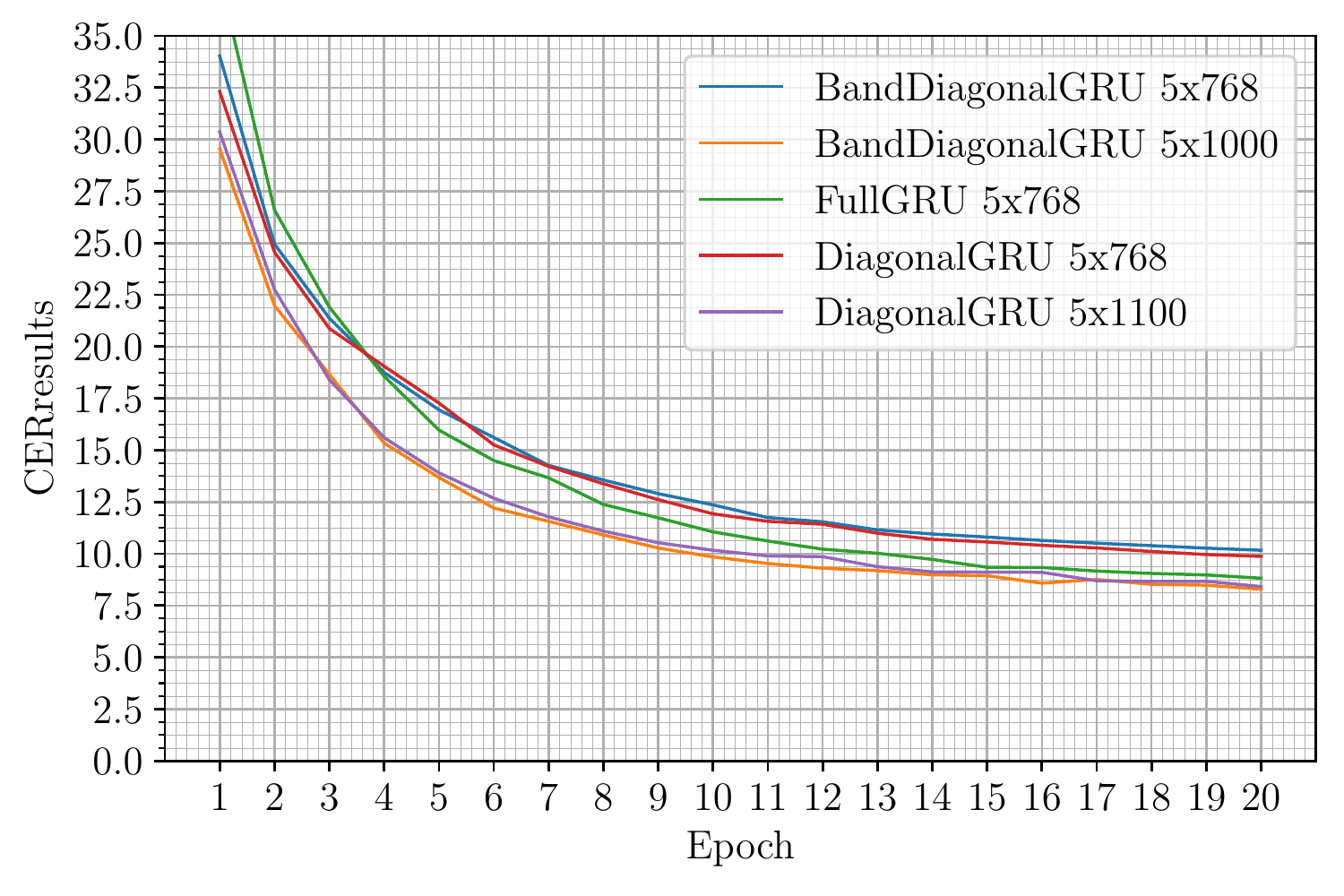}%
		\caption{Character Error Rate (CER) on VCTK. Diagonal-GRU and Band-GRU perform worse than full GRU, but allow larger network sizes for the same budget. These larger networks perform better.}
		\label{f:BandRNN_CER}
	\end{figure}

	A comparison of the character error rate (CER) during training is shown in Figure~\ref{f:BandRNN_CER}. We observe that the larger sparse networks converge faster and achieve lower errors than the fully connected network.
	
	The results are shown in Table~\ref{t:vctk}. DiagonalGRU and BandGRU perform worse than the full GRU by a relatively significant margin. However, the number of recurrent weights in DiagonalGRU is almost 400x lower, in addition to the parallellization opportunity (see Section \ref{s:speedupOpportunity}), so this is still a good option for resource-constrained applications.
	
	With more neurons in the locally connected networks, both BandGRU and DiagonalGRU outperform Full GRU, while still having a much more efficient recurrent transformation and \textit{200x less recurrent weights} for DiagonalGRU.
	
	\begin{table}
		\caption{Speech Recognition on VCTK. For the same number of neurons Full GRU outperforms DiagonalGRU, but larger structurally sparse networks perform better with lower cost.}
		\label{t:vctk}
		\centering
		\begin{tabular}{lccc}
			\toprule
			\textbf{Model} & \textbf{Test CER} &  \multicolumn{2}{c}{\textbf{RNN Weights}} \\
			\cmidrule(lr){3-4}
			& (lower is better)& Total & Recurrent \\ 
			\midrule
			Full GRU $5\times768$ 			& 8.57 & 3.82\e{7} & 1.77\e{7} \\ 
			DiagonalGRU $5\times768, C=1$ 	& 9.69 & 2.06\e{7} & 4.61\e{4} \\ 
			BandGRU $5\times768, C=103$ 	& 9.87 & 2.39\e{7} & 3.41\e{6} \\ 
			DiagonalGRU $5\times1100, C=1$ 	& 8.21 & 3.81\e{7} & 6.60\e{4} \\ 
			BandGRU $5\times1000, C=201$ 	& 8.08 & 3.80\e{7} & 5.76\e{6} \\ 
			\bottomrule
		\end{tabular}
	\end{table}
	
	\subsection{Video Action Recognition: UCF101}\label{s:experimentsUCF101}
	
	Another important category for sequential data analysis is the processing of videos. We use video action recognition as a representative testbed to evaluate BandRNN and DiagonalRNN. Our experiment is performed on the public dataset UCF101~\cite{ucf101}, which contains 101 action classes and 13,320 videos. We follow the two-stream method~\cite{two-stream} to use two separate CNNs on the spatial (RGB) and temporal (optical flow) streams. CNNs of each stream in our evaluations are combined with RNNs. ResNet50~\cite{resnet} pre-trained on ImageNet is used as the backbone CNN. Following PreRNN~\cite{prernn}, we transform the last \texttt{conv} layer of ResNet50 into a recurrent layer and then apply BandRNN and DiagonalRNN. We use the standard experimental setting and report results on the first split.  
	
	Table~\ref{t:ucf101} shows the video action recognition accuracy and number of recurrent weights of different networks. DiagonalRNN and BandRNN coupled with PreRNN achieve comparable accuracy as full PreRNN and perform better than traditional RNNs, while using over 10x less recurrent weights, in particular for the combination of DiagonalRNN and PreRNN-SIH which requires about \textit{100x less recurrent weights} and achieves the same accuracy. 
	
	\begin{table}[t]
		\caption{Comparison of video action recognition accuracy (\%) and number of recurrent weights for different recurrent networks on the UCF101 dataset.}
		\label{t:ucf101}
		\small 
		\centering
		\begin{tabular}{lcccc}
			\toprule
			Model Connectivity & PreLSTM & PreLSTM-SIH & PreGRU & PreGRU-SIH \\
			\midrule
			Full RNN         & $93.2 (2.02 \times 10^7)$ & $93.5 (1.70 \times 10^7)$ & $93.7 (1.49 \times 10^7)$ & $93.3 (1.28 \times 10^7)$\\
			DiagonalRNN      & $92.7 (3.38 \times 10^6)$ & $92.6 (2.23 \times 10^5)$ & $92.8 (2.32 \times 10^6)$ & $92.5 (2.19 \times 10^5)$ \\
			BandRNN, $C=21$  & $92.8 (3.54 \times 10^6)$ & $93.0 (3.87 \times 10^5)$ & $92.8 (2.45 \times 10^6)$ & $92.5 (3.42 \times 10^5)$ \\
			BandRNN, $C=101$ & $92.9 (4.19 \times 10^6)$ & $92.7 (1.03 \times 10^6)$ & $92.8 (2.93 \times 10^6)$ & $92.3 (8.26 \times 10^5)$ \\
			BandRNN, $C=201$ & $93.2 (4.98 \times 10^6)$ & $93.5 (1.82 \times 10^6)$ & $92.9 (3.52 \times 10^6)$ & $92.6 (1.42 \times 10^6)$ \\
			\midrule
			Traditional RNN & \multicolumn{2}{c}{LSTM: $92.5 (2.02 \times 10^7)$} & \multicolumn{2}{c}{GRU: $92.2 (1.49 \times 10^7)$} \\
			\bottomrule
		\end{tabular}
	\end{table}

	\section{Discussion}\label{s:discussion}
	
	Our experiments on the tasks of language modeling, speech and phoneme recognition, and action recognition as well as work by other authors \cite{demeester2018,gray2017blocksparsegpu,kalchbrenner2018,narang2017block} show that structurally sparse networks can match and improve performance of fully connected networks, while reducing the cost significantly.
	For the same number of neurons, sparse RNNs, especially DiagonalRNNs, are much more cost-effective than fully connected RNNs. Alternatively, for the same parameter or compute budget sparse RNNs can be larger, thereby increasing performance. 
	
	Future work could involve optimizing the elementwise recurrent computations, as well as integrating DiagonalRNN with work by \citet{zhu2018persistent} which stores weights fully on-chip to achieve even larger speedups. 
	
	\section{Conclusion}\label{s:comclusion}
	We presented a simple recurrent architecture with high degrees of structured sparsity in the recurrent weight matrix, allowing for large speedups and model size reduction during both training and inference. The models were evaluated on several challenging sequence modeling tasks, and the structurally sparse networks were found to perform as well or better than fully-connected networks in several tasks.
	
	DiagonalRNNs provide straightforward parallellization opportunities, and reduce the cost of the recurrent transformation by a large factor, removing the sequential bottleneck which limits speed of fully connected RNNs on parallel hardware, and greatly reduce the network's memory footprint.
	
	These results are extremely encouraging for further research and practical adoption of structural sparsity in real-world applications, opening the way both to ever-larger neural network sizes and to adoption of high-performance RNNs on low-power and low-cost devices. 
	
%	\clearpage
%	\medskip
%	\small
%	\bibliographystyle{plainnat}
%	\bibliography{references}

\end{document}